\documentclass{19i-2094}

\usepackage{graphicx}
\usepackage{wrapfig}
\usepackage[T1]{fontenc}

\usepackage{amsmath}

\begin{document}

\setlength{\pdfpageheight}{\paperheight}
\setlength{\pdfpagewidth}{\paperwidth}

\conferenceinfo{CONF 'yy}{Month d--d, 20yy, City, ST, Country}
\copyrightyear{20yy}
\copyrightdata{978-1-nnnn-nnnn-n/yy/mm}
\doi{nnnnnnn.nnnnnnn}




\titlebanner{banner above paper title}        

\title{Semantic Parsing to Manipulate Relational Database For a
Management System}
\subtitle{}

\authorinfo{Muhammad Hamzah Mushtaq}
           {FAST National University of Computers and Emerging Sciences}
           {19I-2094}

\maketitle

\begin{abstract}

Chatbots and AI assistants have claimed their importance in today’s life. The main reason behind adopting this technology is to connect with the user, understand their requirements, and fulfill them. This has been achieved but at the cost of heavy training data and complex learning models. This work is carried out proposes a simple algorithm, a model which can be implemented in different fields each with its own work scope. The proposed model converts human language text to computer-understandable SQL queries. The model requires data only related to the specific field, saving data space. This model performs linear computation hence solving the computational complexity. This work also defines the stages where a new methodology is implemented and what previous method was adopted to fulfill the requirement at that stage. Two datasets available online will be used in this work, the ATIS dataset, and WikiSQL. This work compares the computation time among the 2 datasets and also compares the accuracy of both. This paper works over basic Natural language processing tasks like semantic parsing, NER, parts of speech and tends to achieve results through these simple methods.
\end{abstract}


\section{Problem statement}
To generate a generic model for low end individual systems which when provided with any limited database would convert Human language to respected database query efficiently.

\section{Introduction}
Many companies have their own customer support department. Sometimes, users queries do get overbooked and there are not enough staff to handle those queries in time. Instead of employing more workforce, having a smart voice query and result system would save more time and money. More importantly, in a low end system which doesn't have a huge processing power and memory and which needs to perform operations just related to its field, it is hard for it to compute such huge algorithms that today's algorithms require. Training models, requiring loads of training and testing data first and dependency over multiple computers/nodes for heavy processing. It is true that these high tech solution make our life easier but just for solving and achieving bigger goals. In order to work in a limited scope environment, one doesn't need such amount of data and dependencies. This paper aims to provide a solution for such limited scope operations. For this work purpose, we refer to a hotel room management system which only needs to communicate data and information regarding hotel and its vacant rooms and availability.
In the race of developing high end generic solutions, we have lost vision and though of achievements that could come by individually capturing each field. Same is the case for semantic parsing in Speech recognition systems.
 Here, we are talking about a smart system which understands human language queries through their voice and interprets their question and gives required answer in human language.
For example, as DDL queries are simple, the support assistant just needs to view the data from database and reveal it to user. If a smart user assistant handles these queries, the actual support assistants could get more time in handling users with more difficult queries and issues.
This implies the importance of voice assistants which entertain users through directly communicating with database. Henceforth, there must be a channel in which users natural language is translated to query language for computers to understand and interpret. Traditional approach revolves around training over huge amount of data which could fail when out of training set data is occurred. Therefore, the motivation behind the research is to develop a solution which is independent of training set and computes efficient result over some set of rules.

\begin{figure*}[t]
\begin{center}
\includegraphics[width=3.5in]{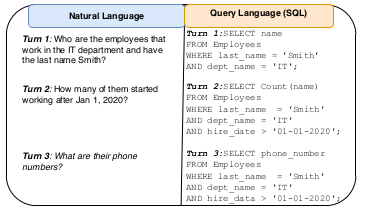}
  \caption{Sample for natural language and related SQL extracted from \cite{colas:cgo01}. In multi turns, these previous models save the query structure and re-query the database. Proposed model adopts different technique for multi turn questions by querying from previous results directly.}
\end{center}
\end{figure*}

\subsection{Background}
Understanding human language by converting it into machine understandable form and then retrieving information back is a tedious task. In this path the main big concept is of Semantic parsing responsible for converting human language to query structure. It requires multiple functions like tokenization, semantic analysis, parsing and passing through language models. If we talk particularly for a specific area of focus, a specific field of work where we need to implement our semantic parsing of human language, then much less data is needed for training purposes. Moreover, information extraction becomes more easy when there is just a single database involved. Instead of training models on various keywords and sentences, just a set of main keywords and there replacement Query syntax are required. This leads to less memory and time consumption. Also, situation specific data can be processed easily.\\

If we generalize the overall approaches used globally for the NLP tasks,\cite{pra:cgo08} we say there are three categories namely:\\
\begin{itemize}
  \item Symbolic Approach or Rule-Based Approach
  \item Empirical Approach or Corpus-Based Approach
  \item Connectionist Approach or Using Neural Networks
\end{itemize}

This research on makes use of both the rule based and corpus based machine learning approach. This is made possible by utilizing the text-to-sql dataset. This dataset itself contains multiple fields databases, mostly with limited database table usage. This is highly likely to compensate in this research since the aim is to manipulate relational database for single scope databases. Secondly, the template that would assist in refining the query consist of a classification model that classifies the table names and column names to the most accurate within the dataset. Researc SPYDER carried out clearly outlines the specifications of multiple text-to-sql datasets. This research doesn't consider dataset Spyder since it does contain multiple database with multiple tables but with different scopes which is not the aim this research.\\ 
The original working flow for the research was that a black box text to sql generator would be used so as to provide a dummy or body of respected sql. This would save time. But translating through model would saved time and complexity.
This research focuses generating and utilizing a generic Query Language template which assists in designing a query, and also how to use already processed data for multi turn dialogues. For template generation, NER will be thoroughly used. We will learn the implications of NER over query generation. Multiple datasets are available as given in {cath:cgo02}. We will be using the structured database of many of these datasets instead of their Natural language to query language translation. We will compare our generated queries with that already given in dataset and distinguish the correctness of our methodology. An example for comparison is shown below. We compare the total number of statements that are translated to query and how many of them were correctly translated. Since the true translations are given within the respective datasets, we can match our results and get the comparison.

\begin{figure*}[h!]
\begin{center}
\includegraphics[width=5in]{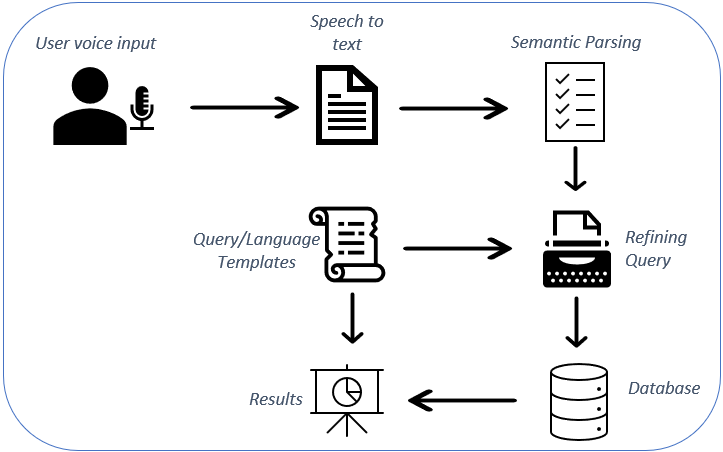}
  \caption{Control flow logic of the being implemented model}
\end{center}
\end{figure*}

\section{Related work}
A lot of work has been done over voice assistants and chat bots. Mainly working over Open domain question answering (QA).\cite{colas:cgo01} For this purpose, a huge amount of data corpus is required and from various sources. Traditional approach requires training state of the art models which utilize this data corpus and learn user behavior over voice commands.

The research \cite{colas:cgo01} also shows the use of domain ontology triples which carry this format. <object, relation, property>. This part gives easy understanding for generating a better query. The research uses Lexicons to match the instances from natural language to query language. This is a very effective technique which acts as a relay, a bridge between normal sentences and Query sentences. More further, the query templates are directly grouped in particular predicate group which defines its structure and mechanism, either DDl or DML.Earlier models were trained on pre defined text and supposed result. But the accuracy decreased when a new text or sentence appeared which was not trained in training set. \\

Previous work also included a dialog based structure \cite{gur:cgo06} in which the user itself intervened for the better and accurate results of query, basically refining the query as the questions asked by the model from user. This research excludes such intervention for the main cause that user just has to input his voice command and get the results, the model must be properly trained to streamline and refine the sql query itself.\\
Many methods have been implemented ranging from rule based to neural network approach as discussed in paper \cite{kim:cgo09}. Each has its own limitations. Neural network implementation from translating human language to sql gives successful results with high accuracy but at higher cost. They require huge umber of parameters and huge training corpus. Authors have also highlighted that since each model is trained and tested on different dataset, there accuracy's differ because of the accuracy and correctness of dataset itself. Authors also state that the very famous WikiSQL is based on a simple syntax pattern including SELECT <aggregation function> <column name>
FROM T [ WHERE <column name> <operator> <constant value> (and <column name> <operator> <constant value>)* ],
where T is a given single table. With this we know that it doesn't support grouping, ordering, join or nested groups. A well distinction of WikiSQL along with other benchmark datasets is defined in figure 4 extracted from another paper. \\
Human generated dataset have been evaluated efficiently in \cite{cath:cgo02}. It identified that human generated datasets lacked some of the properties needed in large-scale query sets.Up till now, work was being carried out keeping a generic concept in mind. By generic we mean covering all aspects of information, all areas of research and multiple topics and subjects that must be learned by model. Hence a broader approach was necessary. 

\section{Implementation}
Our first approach over this research is to convert human voice to text. The text is passed through semantic parsing to develop the initial syntax for our query. We proposed a rule based approach in which the initially developed syntax is refined through the help of predefined set to Query Language template or rules. In this way, the computer gets to learn the correct form.Then the query is run over the database to fetch results. These queries can be of any type; Data Definition Language (DDL) or Data Manipulation Language(DML). This process is to be adopted for single-turn dialogue, where there is just one single operation. The second change that is to implemented is using already gathered results for multi turn dialogues.\\

Instead of redesigning the query and gather the similar results over it, we propose a set of grammar and rule which define when to run query over already gathered results and when to query the database. Also, this research will put in use NER. This will be helpful when matching parsed query with domain ontology tuples or predefined templates.

\subsection{Voice to text}
This is the first and initial part of our model in which voice input is converted to text. The main work of this model resides in efficiently processing and computing users text command in sql query form, hence, we will be using third party API for just converting human voice to text. For best results we would use google's speech to text api. We convert human voice and then perform further process on it.Also, our model as it is trained over recognizing English verbs or pronouns which would declare what type of operation to perform, the model is prone to some errors because of its dependency on generated language.

\subsection{Semantic parsing}
The generated English sentence is broken down here into parts of speech and object/ subject distinction. Hence each word carries its attributes further for query generation. Major use here will be of Named Entity Recognition NER which will tag the words as names, location, time etc. The reason for adopting this process is easily match the data with our already defined template. That query template would convey us the refined SQL based on some predefined set of conditions. For example, when certain words fall into certain category, a different SQL syntax will be use.\\
 In parallel, a model is trained which generates a sql body for the next method to work on. The model is trained over input sentences and predict what type of sql query would be used, for example a select statement or a insert or delete statement. This model would decide if to form DDL or DML.

\subsection{Refining Query}
\begin{figure}[b]
\includegraphics[width=3in]{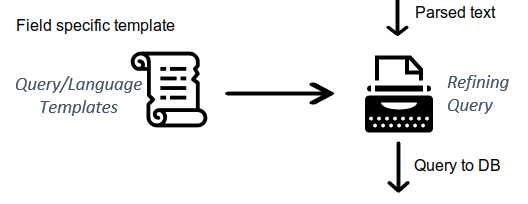}
  \caption{Refining query includes the field specific template and parsed text as input}
\end{figure}
The already defined SQL template also tells us whether to execute our statement on previously generated result or re-query over database. This would help us shortening the query and saving a lot of time which would be wasted in conditional query over database.\\
Secondly, this is the part which is different from most of the modern techniques. Since our goal is to develop a model for low end systems running individually for a specific field and purpose, our goal to convert language to SQL query becomes easy. When the field and scope is narrowed down, so is the data needed to identity human language meaning.\\

\begin{figure*}[h!]
\begin{center}
\includegraphics[width=5in]{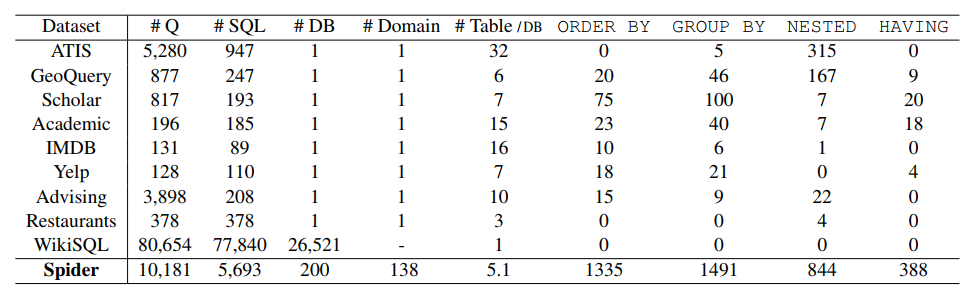}
  \caption{\cite{yu:cgo07}Datasets comparisons}
\end{center}
\end{figure*}

The field specific template would include precise NER subjects and there respective SQL query mapping, along with the the structure of conveying results in text form.  First column would describe the NER subjects like object, name, place etc, second column would define the assigned query along with the positions of the entities. The third column denotes the structure of the results that would be displayed. For example if a SELECT statement with count is generated then the resultant query would start with "There are <count> <object>".\\

\begin{table}
\begin{tabular}{ |p{2cm}|p{2cm}|p{3cm}|  }
 \hline
Satement & SQL mapping & Result statement\\
 \hline
 How many rooms are available? & SELECT COUNT(ID) FROM <OBJECT> & THERE ARE (COUNT) <OBJECT> AVAILABLE\\ 
\hline
 ...&   ...&...\\
 \hline

\end{tabular}
\caption {Table defining the intermediate structure after the model translates text into a query.}
\end {table}
\section{Evaluation and Results}
This study tends to identify the behavior of voice assistant over Rule based approach during semantic parsing, rather than pre-learned models which require pre-processed data in a huge amount. Also, the use and dependency of NER over query generation will be studied. With this research model, a new Human Language translation system is generated which would ideally just get the connected Database information from the user and create queries likewise.The initial idea has been highlighted in this paper and the drawbacks and setbacks of previous approaches over low core database systems have been highlighted.\\

Code available at  : \textit {https://github.com/hamzah1947/textToSQL.git}

\subsection{Data}
Since data and naming convention for different databases and fields differ, this research works on achieving a single trained model which when connected to any database generates efficient queries for that database respectively. text-to-SQL dataset contains multiple databases respected to different fields. It is most suitable in integrating and testing the current model. The only dependencies of proposed model are for the initial text to sql model which is trained over pre-defined semantics translations. Secondly, the model that classifies best table name and column train on the text-to-sql dataset as whole. Once the table is determined, it is easy for model to drill down to that database table and classify only among its columns.

\begin{table}[h!]
\begin{center}
 \begin{tabular}{| c | c | c | c |} 
 \hline
 Dataset & Statements executed & No. of correct results \\ [0.5ex] 
 \hline\hline
 1 & 50 & 40\\ 
 \hline
 2 & 40 & 38  \\
 \hline
 3 & 40 & 40\\
 \hline

\end{tabular}
\end{center}
\caption{Table comparing correctly generated queries among different data-sets}
\label{table:1}
\end{table}

\subsection{Model training}
In the proposed methodology we also touch multi turn QA in our system. This is done through the help of our QL template which directs from where to query, either from database or previous results. Process are to be implemented over both data-sets and the results will be displayed in this manner.\\
The model consists of 2 SVM models. First one trained to predict SQL syntax from human language and the second model classifies table name and column names.

\begin{table}[h!]
\begin{center}
 \begin{tabular}{| c | c | c |} 
 \hline
 Dataset & Average Computation time& Multi-turn\\ [0.5ex] 
 \hline
 1 & 4 & Yes\\ 
 \hline
 2 & 10 & Yes  \\
 \hline

\end{tabular}
\end{center}
\caption{Table comparing correctly generated queries among different data-sets}
\label{table:2}
\end{table}

\section{Conclusion}
This study proposes a model which best benefits low score database systems which are unable to handle large training data and computation requirements. Once the complete implementation of model, it would be able to integrate with any data and have a competitive accuracy of text to sql and database manipulation with accurate results.\\


\bibliographystyle{abbrvnat}


\end{document}